\documentclass{easychair}
\usepackage{doc}
\title{Project proposal: A modular reinforcement learning based automated theorem prover \thanks{This work has been supported by the French government, through the 3IA Côte d'Azur Investments in the Future project managed by the National Research Agency (ANR) with the reference number ANR-19-P3IA-0002}}
\author{Boris Shminke \inst{1}}
\institute{
  Université Côte d'Azur, CNRS, LJAD, France \\
  \email{boris.shminke@univ-cotedazur.fr}
 }
\authorrunning{Shminke}
\titlerunning{Project proposal: A modular reinforcement learning based automated theorem prover}
\begin{document}
\maketitle
\begin{abstract}
We propose to build a reinforcement learning prover of independent components: a deductive system (an environment), the proof state representation (how an agent sees the environment), and an agent training algorithm. To that purpose, we contribute an additional Vampire-based environment to \texttt{gym-saturation} package of OpenAI Gym environments for saturation provers. We demonstrate a prototype of using \texttt{gym-saturation} together with a popular reinforcement learning framework (Ray~\texttt{RLlib}). Finally, we discuss our plans for completing this work in progress to a competitive automated theorem prover.
\end{abstract}
\section{Introduction and related work}
Reinforcement learning (RL) is applied widely in the automated reasoning domain. There are RL-related (including iterating supervised learning algorithms without applying recent RL advances) projects for interactive theorem provers (ITPs) (e.g. HOList~\cite{DBLP:conf/icml/BansalLRSW19} for HOL~Light~\cite{10.1007/978-3-642-03359-9_4}, ASTactic~\cite{DBLP:conf/icml/YangD19} for Coq~\cite{the_coq_development_team_2022_5846982}, or TacticZero~\cite{NEURIPS2021_4dea382d} for HOL4~\cite{10.1007/978-3-540-71067-7_6}) as well as for automated theorem provers (ATPs) (e.g. Deepire~\cite{10.1007/978-3-030-79876-5_31} for Vampire~\cite{10.1007/978-3-642-39799-8_1}, ENIGMA~\cite{10.1007/978-3-030-29436-6_29} for E~\cite{10.1007/978-3-319-62075-6_20}, or rlCoP~\cite{NEURIPS2018_55acf853} for leanCoP~\cite{OTTEN2003139}). Despite the variety of solutions and ideas, we are not aware of cases of significant code reuse between such projects.

We envision a prover capable of learning from its experience and composed of pluggable building blocks. We hope that such architecture could promote faster experimentation and easier flow of ideas between different projects for everyone's progress. For an RL-based prover, we identify at least three types of modules. They are a deductive system (an environment), a proof state representation (how an agent sees it), and an agent training algorithm.

When choosing whether to learn to guide an ITP or an ATP, we prefer the latter since ATPs can be relatively easy compared as black boxes~\cite{DBLP:journals/aicom/Sutcliffe21} in contrast to RL guided ITPs, which often come with their distinctive benchmarks.

Among ATPs, one can consider saturation provers less suitable for the RL (e.g., see design considerations from ~\cite{10.1007/978-3-030-29007-8_3}), but several existing projects (like ENIGMA, Deepire or TRAIL~\cite{Crouse_Abdelaziz_Makni_Whitehead_Cornelio_Kapanipathi_Srinivas_Thost_Witbrock_Fokoue_2021}) show encouraging results. Keeping that in mind, we decided to concentrate on guiding clause selection in the saturation algorithm by RL.

Inspired by HOList, CoqGym (from ASTactic) and \texttt{lean-gym}~\cite{DBLP:journals/corr/abs-2202-01344}, we have created \texttt{gym-saturation}~\cite{Shminke2022} --- an OpenAI~Gym~\cite{DBLP:journals/corr/BrockmanCPSSTZ16} environment for training RL agents to prove theorems in clausal normal form (CNF) of the Thousands of Problems for Theorem Provers (TPTP) library~\cite{DBLP:journals/jar/Sutcliffe17} language.

\section{Recent work in progress}
Contemporary RL training algorithms are notorious for the number of details that can differ from one implementation to another~\cite{Henderson_Islam_Bachman_Pineau_Precup_Meger_2018}. To eliminate the risk of abandoning an RL algorithm as unsuitable for guiding an ATP only because of flaws in our implementation of it, we plan to rely on existing RL frameworks containing tested implementations of well-known baselines. As a starting point, we have chosen Ray~\texttt{RLlib}~\cite{DBLP:journals/corr/abs-1712-09381} as a library claiming both deep learning (DL) framework independence and extendability. Similar solutions like Tensorflow~Agents~\cite{DBLP:journals/corr/abs-1709-02878} or Catalyst.RL~\cite{DBLP:journals/corr/abs-1903-00027} tend to support only one DL framework, which we wanted to avoid for greater generality.

In contrast to CoqGym and others, \texttt{gym-saturation} is not only a `gym' in some general sense, but it implements the standard OpenAI Gym API. It makes it easier to integrate with libraries like Ray~\texttt{RLlib}. We contribute\footnote{https://github.com/inpefess/basic-rl-prover} a prototype of such integration. Even together with some domain related patches, the prototype remains a lightweight collection of wrappers around standard \texttt{RLlib} classes, taking only around 300 lines of Python code.

Since we postulated interchangeability of modules, we added a Vampire-based environment to \texttt{gym-saturation} (see the project page\footnote{https://pypi.org/project/gym-saturation/} for more details) in addition to the already existing naïve implementation of a saturation loop. Despite a different backend, one can plug a new environment into the prover prototype without additional edits of RL related code.

\paragraph{Similar systems for connection tableaux} There exists a \texttt{FLoP} (Finding Longer Proofs) project~\cite{FLoP} which implements a \texttt{ProofEnv} OpenAI~Gym environment for a connection tableaux calculus, which can guide two different provers (\texttt{leanCoP} and its \texttt{OCaml} reimplementation \texttt{fCoP}~\cite{fCoP}). \texttt{FLoP} shares many architectural features with our work, and we plan to test its approaches in saturation provers setting.

\section{Prototype implementation details}
Since this research is still in an early stage, we don't report any conclusive results of its performance, only describing the architecture. A prototype prover has two main parts: \texttt{gym-saturation} as an environment and a patched DQN~\cite{rainbow} implementation from Ray~\texttt{RLlib} training an agent. \textbf{An episode} starts with the environment reset. On environment reset, a random TPTP problem from a training subset is loaded, transformed to the CNF, and becomes a proof state. After an agent makes an \textbf{action} (selects a clause), the episode can stop for three reasons: a given clause is empty (refutation proof found, the \textbf{reward} is 1.0; in other cases, it's 0.0), we reach the step limit (a soft timeout), we reach the maximal number of clauses in the proof state (a soft memory limit). Only episodes with a positive final reward go to the \textbf{memory buffer}. Before storing in the buffer, the reward is spread evenly between the clauses from the proof (others remain zero). A memory buffer can contain the same proof for the same problem twice or different proofs (maybe of different lengths) for one problem. A \textbf{training} batch can contain steps from different episodes (and thus different initial environment states). We sample the memory buffer with higher weights for more recent episodes.

\section{Future plans and discussion}
In the prototype, we represent each clause in a proof state only by its size and order number, applying a logistic regression as a Q-value function. We will need an elaborate feature extraction procedure to complete this oversimplified model to a competitive ATP. We plan to use graph neural networks similar to those used for lazyCoP~\cite{DBLP:conf/tableaux/RawsonR21} and then compare and combine them with the graph representation of clause lineage pioneered by Deepire. We also plan to test training algorithm interchangeability by using IMPALA~\cite{pmlr-v80-espeholt18a} and Ape-X~\cite{horgan2018distributed} in addition to DQN.

A finished project will have to address many different problems. Here we list several obvious ones.
\paragraph{Delayed reward} One of the well-known peculiarities of an ATP is the fact that a reward can be assigned only after proof is found, which can take a large number of steps in an RL episode. To make an agent learn to discern good steps, one has to spread the final reward to all the steps in a finished trajectory. A typical solution is to post-process a trajectory by assigning positive advantage values only to the steps encountered in a proof, and negative (or zero) values to all the rest. Here one can argue in favour of both higher values for longer proofs (since the ability to produce longer proofs is desirable) and higher values for shorter ones (since more concise proofs for simpler problems are preferable to verbose ones which in turn could help to find longer proofs otherwise unreachable because of time and memory constraints). A contrarian approach is to assign positive advantage values for all the steps in a trajectory on which proof was found, and non-positive to all the steps from trajectories finished because of the resource limitations. Such an approach works well, for example, in the Atari Pong game, where it's practically impossible to judge which action led to a goal.
\paragraph{Sparse positive reward} Another well-known problem of applying RL to ATPs is related to the fact that even sub-human performance still seems out of reach. The majority of proof attempts finish without proof found. Discarding failed episodes seems too wasteful, although obvious as a first attempt. An opposite solution (assigning non-positive advantage values to all failed episodes) makes the training dataset too imbalanced. One possible solution to this is to use replay buffers and sample from them balanced train batches. This explains why we decided not to neglect DQN despite its known limitations when compared to on-policy algorithms like PPO~\cite{Schulman2017ProximalPO}.
\paragraph{Multiple proofs} Many problems have multiple possible proofs, equivalent in some sense or not. An agent will have to decide which proofs are preferable to replicate. Again, replay buffers can be used for that. Ranking proofs can be based on their length or other important properties (reuse of previously proved lemmata, using only a selected subset of deduction rules or tactics etc)
\paragraph{High environment's inhomogeneity} Some problems are inherently harder than others and can belong to areas of mathematics not connected in a given formalization. Curriculum learning~\cite{CurriculumLearning} or at least limiting the training scope to a reasonable subset of the TPTP library will be needed.
\paragraph{State representation} Usually, contemporary RL algorithms expect the observed state to have a form of a vector. Representing logic formulae as such is an active domain of research. We plan to try both logic-specific approaches like~\cite{VectorRepresentations} and general abstract syntax tree encoding models like \texttt{code2vec}~\cite{alon2019code2vec} or \texttt{ast2vec}~\cite{Paassen2022}.
\label{sect:bib}
\bibliographystyle{plain}
\bibliography{aitp2022}
\end{document}